\documentclass[letterpaper, 10 pt, conference]{ieeeconf}  
\usepackage{url}
\usepackage{hyperref}

\usepackage{graphicx}
\usepackage{enumerate}
\IEEEoverridecommandlockouts                              

\title{\LARGE \bf
Deep Learning on Home Drone: Searching for the Optimal Architecture}

\author{Alaa Maalouf$^{*1}$, Yotam Gurfinkel$^{*1}$, Barak Diker$^{1}$, Oren Gal$^{1}$, Daniela Rus$^2$, and Dan Feldman$^{1}$ 
\thanks{$^{1}$Robotics and Big Data Labs, Department of Computer Science, University of Haifa. $^{2}$CSAIL, MIT}
\thanks{$^*$Equal contributions}
\thanks{Corresponding author email: {\tt\small alaamalouf12@gmail.com}}
}

\begin{document}

\maketitle
\thispagestyle{empty}
\pagestyle{empty}

\begin{abstract}
We suggest the first system that runs real-time semantic segmentation via deep learning on a weak micro-computer such as the Raspberry Pi Zero v2 (whose price was \$15) attached to a toy-drone. In particular, since the Raspberry Pi weighs less than $16$ grams, and its size is half of a credit card, we could easily attach it to the common commercial DJI Tello toy-drone (<\$100, <90 grams, 98 $\times$ 92.5$ \times$ 41 mm). The result is an autonomous drone (no laptop nor human in the loop) that can detect and classify objects in real-time from a video stream of an on-board monocular RGB camera (no GPS or LIDAR sensors).  The companion videos demonstrate how this Tello drone scans the lab for people (e.g. for the use of firefighters or security forces) and for an empty parking slot outside the lab.

Existing deep learning solutions are either much too slow for real-time computation on such IoT devices, or provide results of impractical quality. 
Our main challenge was to design a system that takes the best of all worlds among numerous combinations of networks, deep learning platforms/frameworks, compression techniques, and compression ratios. To this end, we provide an efficient searching algorithm that aims to find the optimal combination which results in the best tradeoff between the network running time and its accuracy/performance.

\end{abstract}

\section{BACKGROUND}

   
Deep learning advancements in the previous decade have sparked a flood of research into the use of deep artificial neural networks in robotic systems. 
The main drawback is that existing deep learning solutions usually require powerful machines, such as strong servers and graphics (GPU) cards. This is a problem when it comes to small robots due to the following challenges:
\begin{enumerate}[(i)]
    \item The weight of the computation machine might be too much, especially for ``flying robots" such as nano-drones. 
    \item  The large amount of required energy to operate powerful computers and graphics translates into large (heavy) batteries and a shorter lifetime between charging, especially for moving robots.
    \item The price in terms of money for these computations machines, is relatively high, especially when it comes to low-cost or Do-It-Yourself (DIY) robots.
\end{enumerate}

\begin{figure}[t]
      \centering
      \includegraphics[scale=0.15]{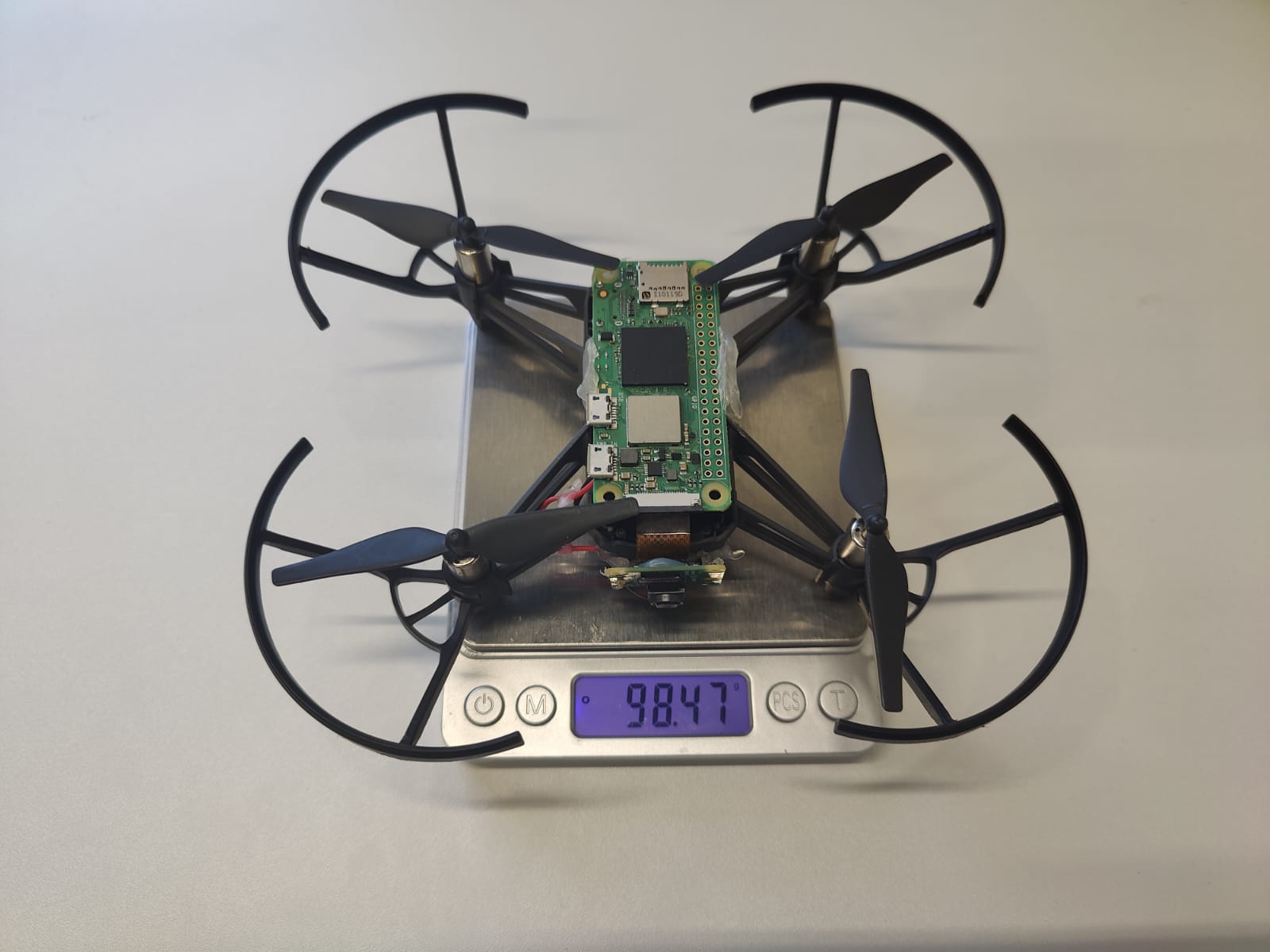}
      \caption{A DJI's Tello drone equipped with our Raspberry PI Zero ver 2 micro-computer, connected to an RGB camera. Our deep learning-based systems are executed on this hardware. }
      \label{fig:tello}
   \end{figure}

It seems that most robotics applications that need to run deep learning under such constraints use the powerful Nano-Jetson machine of NVidia~\cite{cass2020nvidia,suzen2020benchmark}, which was used in many robotic systems in recent years such as~\cite{jeon2021run,suzen2020benchmark,9636518,9812220}. However, while much more powerful, NVIDIA's nano-Jeton (nJeston) is far behind micro-computers such as the recent Raspberry Pi Zero Ver 2 (RPI0), with respect to constraints (i)--(iii) above:  
(i) The nJetson weighs about $100$ grams, and thus cannot be carried by a small low-cost drone as in this paper. Such a drone can easily carry the RPI0  which weighs $16$ grams, whereas the whole system weights less than $100$ grams (the RPI0  and the drone); see Fig.~\ref{fig:tello},
(ii) The energy consumption of nJetson is $\sim5$ watts at idle and $\sim10$ watts under stress, which is about $18\times$ times more than the RPI0  which consumes $\sim0.28$ watts at idle and $\sim0.58$ watts under stress, and (iii) Our nJetson costs $\sim \$50$, compared to the RPI0  that we used in this paper which costs~$\sim \$15$. 

For comparison, the Tello drone (as in Fig.~\ref{fig:tello}) could not even carry the nJetson, not to mention its huge battery. A useful feature that we discovered regarding the RPI0  is that, due to its low voltage, \textbf{it can use the same battery as its carrying drone} (Tello in this paper). On the contrary, even a more powerful version of RPI (such as RPI3) needs an additional battery which that drone can barely lift stably and drains its battery more quickly.
\paragraph{Vision}
   
\begin{figure*}[t]
      \centering
      \includegraphics[scale=0.4]{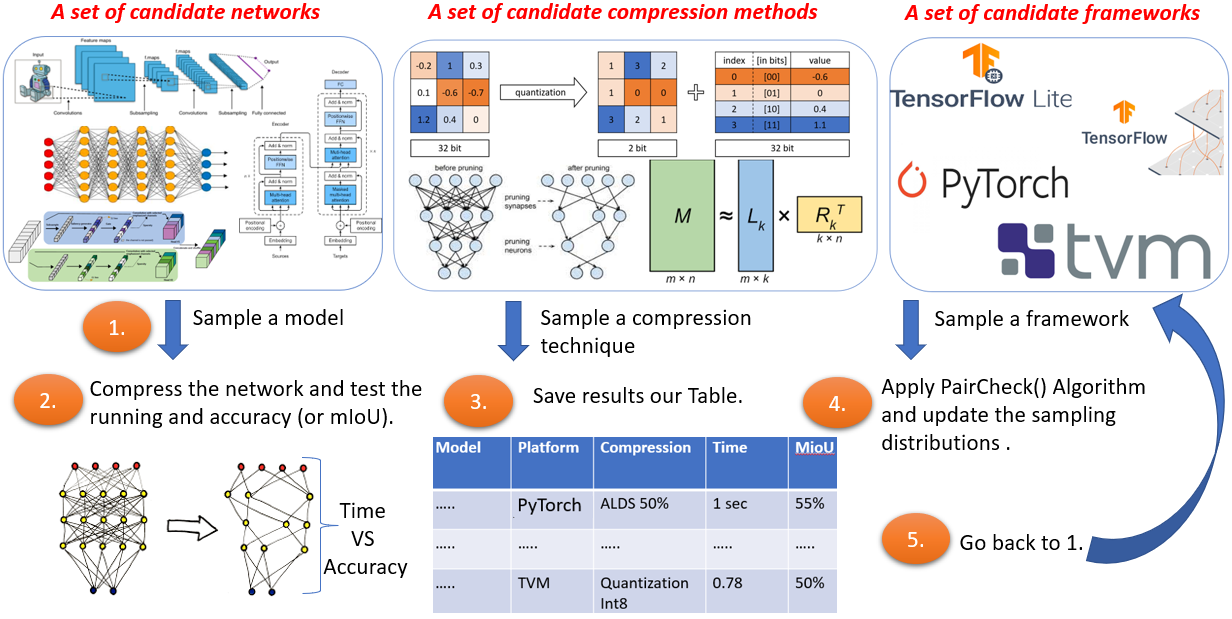}
      \caption{Searching for the optimal network: (1) sample a combination of network, framework, and compression technique (with its hyper-parameters), (2) Compress the sampled network on the sampled framework via the sampled compression method, and check the running-time and error on the required machine (RPI0  in our case), (3) Update the table of results, and (4) Run \textsc{PairChecker} to update the sampling options, and go back to (1).}
      \label{fig:flow}
   \end{figure*}
The motivation for this paper was to turn a remote-controlled commercial toy-drone into an autonomous drone that is able to run deep neural networks efficiently, and thus gain all of their benefits for indoor applications.  All of this should be done using only a single RGB camera and an on-board micro-computer, without a laptop or human in the loop. Firefighters may use it for searching for people, such as survivors in a building on fire, the police can use it to look for suspicious objects (e.g., in a subway), and the army may use it for rescue or traps. Besides security forces, it may be used in the industry for counting people (e.g. in a mall) for sales optimization or counting objects in a warehouse.

\paragraph{Selected drone} To accomplish this goal, we choose the common commercial Tello DJI drone since it is small enough to be  \textbf{lawful}, \textbf{low-cost}, and \textbf{safe} for indoor applications. The other options were: (1) the Mavic mini-combo~\cite{mavik} which weighs nearly 250 grams, costs $10\times$ times, and was too big for our building, and (2) the Crazyflie~\cite{giernacki2017crazyflie} which is smaller than the Tello and open-sourced but could not carry our mini-computer.

\begin{figure}[t]
      \centering
      \includegraphics[width=8cm, height=2.2cm]{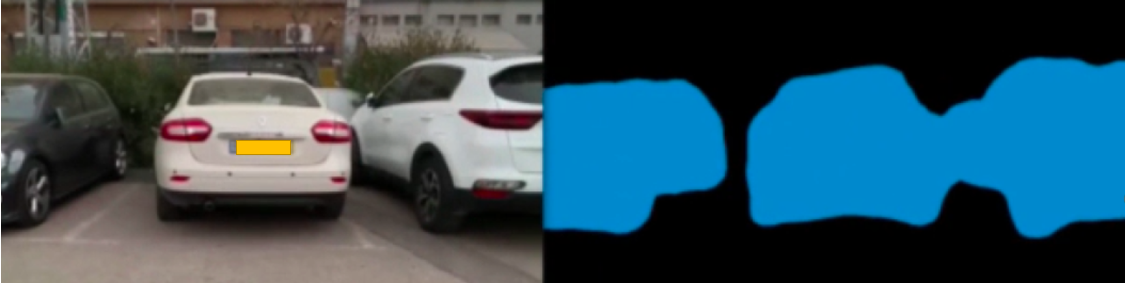}
      \caption{\textbf{Left:} The original video seen by the drone. \textbf{Right:} The semantic segmentation for detecting empty parking spots}
      \label{fig:car}
   \end{figure}

\paragraph{Semantic Segmentation} A common requirement for the mentioned applications above is to associate each pixel of an image (or frame in a video) with a class label  (such as a wall, person, road, table, or car). This task is called \emph{semantic segmentation}~\cite{garcia2017review}; see Fig.~\ref{fig:car} for example. 
Since this task is considered among the hardest in computer vision, most image segmentation systems require a high-power machine or an internet connection to transfer photos to a cloud server that can run huge deep learning models. However, the safe legal toy-done will not be able to carry the high-power machine (as the nJeston above), and the cloud connection may impose extra restrictions on the applications of the semantic segmentation such as (high) extra charges and latency.

Hence, once we are able to efficiently apply semantic segmentation on a small micro-computer placed on a tiny drone, we can safely use our drone for the mentioned security, civil, and commercial applications.



\paragraph{Open problems} While a navigation system was recently suggested for our hardware setup above in~\cite{jubran2022newton}, we could not decide which network is considered an efficient real-time network that can apply semantic segmentation in a sufficiently fast time on our specific hardware. The open problems and main challenges for this paper are:

\begin{enumerate}[(i)]
    \item \textbf{Can we run semantic segmentation in real-time on a weak device such as the Raspberry-Pi Zero v2?}
    \item \textbf{How to find (search for) the desired efficient network?}
    \item \textbf{Can we use it for onboard object/people detection for an indoor drone?}
\end{enumerate}

\subsection{Our Contribution}
We answer these questions affirmably by providing the following results:
\begin{enumerate}[(i)]
    \item \textbf{Searching for the optimal network. } We suggest a searching algorithm to identify efficient networks concerning the trade-off between accuracy and speed. This is by searching for the optimal combination between an efficient deep learning framework (platform, e.g., PyTorch), an efficient and robust deep learning model, and newly suggested compression techniques; See Fig.~\ref{fig:flow} and Section~\ref{seC:choices}. In our case, we test our algorithm on the semantic segmentation task which runs on RPI0. We provide a table that compares a variety of such combinations, towards finding the best fit; See Section~\ref{sec:method} and Table~\ref{table:comparison}.

    \item  \textbf{Low-cost system. }We provide a low-cost (and low energy) system for safely running indoor applications using a small Tello drone that carries RPI0 (as in Fig.~\ref{fig:tello}), which runs deep learning on-board. We demonstrate the uses of this system by implementing and providing videos for the following three semantic segmentation tasks: finding people, identifying an empty parking spot, and detecting objects; see Section~\ref{sec:systems}.

    \item \textbf{Open source code. } Full open source code, of all of the tested and compressed networks, on different deep learning frameworks, with a comparison table between them is  provided\footnote{\href{https://github.com/YotamGurfinkel/DeepLearningOnDevice}{\url{https://github.com/YotamGurfinkel/DeepLearningOnDevice}}} for future research and extensions.
\end{enumerate}


\section{THE SEARCH}\label{sec:method}
Our goal is to suggest a neural network for Semantic Segmentation that would be efficient enough to enable real-time applications on a weak micro-computer such as the RPI0. 
Existing networks have a high quality of segmentation but require strong computation power, much more than our RPI0. The required power is determined by several parameters, such as the network architecture and the framework which was used to implement (e.g., Pytorch).
For example, using the DeepLab v3 network with the backbone of MobileNet V3 implemented above Pytorch to apply semantic segmentation on an input image of $513\times 513\times 3$ (on RPI0) took more than $10$ seconds per frame, which is far too slow for a real-time system or a flying drone. Other networks frameworks combinations may take $30$ or $40$ seconds to run a single frame from the RPI's streamed video.

Due to their quality, instead of retraining a new network from scratch, our strategy was to compress an existing (pre-trained) network. The main challenge was to find a sufficiently good combination among the following parameters, in order to obtain the desired network that should have high quality and fast running time:

\textbf{Type of platform/framework. } There are three common frameworks to implement deep learning applications:  
Pytorch~\cite{paszke2019pytorch}, TensorFlow and its light version TensorFlow Light~\cite{abadi2016tensorflow}, and TVM~\cite{chen2018tvm}. Each of these frameworks may be preferable (in terms of efficiency) in some scenarios, and worst in others. It depends basically on the machine (hardware) architecture that the network will be executed on, the architecture of the network we wish to use, and the number of optimizations done by the users who wrote the code. 

\textbf{The used model.} The network architecture itself determines the number of parameters and the floating point operations that needs to be done, which consequently determines the time that is required to process a single input. Furthermore, some architectures (or layers) might be optimized on some deep learning frameworks more than others.


\textbf{The compression method.} The networks from the previous bullet are sometimes too big to apply a real-time semantic segmentation, so we need to compress them via one of the following existing compression approaches: (i) quantization~\cite{Wu2016,rastegari2016xnor}, where weights are represented using fewer bits, (ii) knowledge distillation~\cite{park2019relational,hinton2015distilling}, where a small network is trained to imitate a large one, (iii) weight and neuron pruning~\cite{Han15,liebenwein2019provable,tukan2022Pruning}, where individual weights or structures of the network are removed, and (iv) low-rank decomposition~\cite{maalouf2021deep,tukan2021no,lebedev2014speeding}, where a tensor representing some layer is decomposed into more than one representing several layers. Each of these techniques has its advantages and disadvantages.

\begin{table*}[t]
\caption{Comparison Table. This table compares (i) the mIoU (mean intersection over union) of applying a semantic segmentation on the Pascal Voc 2012 validation set and (ii) the time of processing a single frame on RPI0, with respect to different networks, compression methods, and deep learning platforms.}\label{table:comparison}
\label{Comparison Table}
\begin{center}
\begin{tabular}{|c|c||c|c|c|}
\hline
\textbf{Network (model)} & \textbf{Framework} & \textbf{Compression} & \textbf{Inference Time [sec]}& \textbf{mIoU} \\
\hline

\multicolumn{5}{|c|}{\textbf{The input image size is  $513\times 513\times3$}}\\
\hline

DeepLabV3-MobileNetV2 & Tensorflow Lite & NA & 3.18 & 67.7\% \\
DeepLabV3-MobileNetV2 & Tensorflow Lite & Quantization-int8 & 1.63 & 61.2\% \\
DeepLabV3-MobileNetV2 & Tensorflow Lite& Depth Multiplier = 0.5 & 1.37 & 60.6\% \\
DeepLabV3-MobileNetV2 & Tensorflow Lite& Depth Multiplier = 0.5 \& Quantization-float16 & 0.79 & 60.6\% \\
DeepLabV3-MobileNetV2  & Tensorflow Lite&  Depth Multiplier = 0.5 \& Quantization-int8 & 0.92 & 54.7\%\\ \hline
LRASPP-MobileNetV3-Large & Apache TVM & NA & 1.02 & 65\%  \\
LRASPP-MobileNetV3-Large & Apache TVM & $32\%$ Compression by ALDS & 0.7 & 55.4\% \\
LRASPP-MobileNetV3-Large & Apache TVM & $45\%$ Compression by ALDS& 0.6 &  56.39\% \\
LRASPP-MobileNetV3-Small & Apache TVM & NA& 0.39 &61\% \\ \hline
DeepLabV3-MobileNetV3-Large & PyTorch & NA & 37 & 67.4\% \\
LRASPP-MobileNetV3-Large & PyTorch & NA & 30 & 65\%  \\
LRASPP-MobileNetV3-Small & Pytorch & NA& 1.25  &61\% \\
\hline

\multicolumn{5}{|c|}{\textbf{The input image size is  $284\times284 \times3$}}\\
\hline
LRASPP-MobileNetV3-Large & Apache TVM & NA & 0.29 & 64.9\%  \\
LRASPP-MobileNetV3-Large & Apache TVM & $32\%$ Compression by ALDS & 0.2 & 53\% \\
LRASPP-MobileNetV3-Large & Apache TVM & $45\%$ Compression by ALDS& 0.18 & 53.7\% \\
LRASPP-MobileNetV3-Small & Apache TVM & NA&  0.1 & 58.48\% \\ \hline
\hline
\end{tabular}
\end{center}
\end{table*}
\subsection{The Choice of Networks and Compression Methods}\label{seC:choices}

 In this section, we described our search space concerning the networks and compression methods.

\textbf{Which Deep Neural Networks to use? }We chose two different Semantic Segmentation heads for our comparison. The first is DeepLabV3~\cite{chen2017rethinking}, a very well-known head for semantic segmentation, which is relatively well known, but (apparently) not ideal for video (streaming). The second head is LRASPP~\cite{howard2019searching}, which is a pretty new architecture for a semantic segmentation head. While DeepLabV3's accuracy is better, LRASPP seemed to be specifically suitable for video, which is very important in our test cases (this is also LRASPP's purpose in its paper). For the backbone, we use two versions of MobileNet V2~\cite{sandler2018mobilenetv2} and V3~\cite{howard2019searching}, which are both known to have a fast inference time on edge devices. 

\textbf{Depth Multiplier. } For the backbone of MobileNet-V2, we can use the parameter ``depth multiplier'' (also known as \emph{width multiplier} in the MobileNet-V2 paper), which decreases the number of filters in each of its layers by its given value. Hence, we can use this as a given compression technique.  We test the network with a depth multiplier of $0.5$ (since its corresponding trained network exists). This makes the network faster, but of course, hurts accuracy (as reported in our comparison table; see Table~\ref{table:comparison}).

\textbf{Which compression methods to use?} Weight pruning generates sparse models (not smaller ones) requiring specialized hardware and software in order to speed up inference. Hence, we aim to avoid it. Neuron pruning usually requires expensive retraining cycles to achieve comparable results to other approaches~\cite{pietron2020retrain}, and knowledge distillation requires a full training procedure. Since we wish to test many networks, these approaches are too time consuming. Hence, we used only efficient methods which approximate the original network efficiently without requiring many retraining cycles or any special hardware. Instead, we use low-rank decomposition and quantization. 

\textbf{The ALDS algorithm. } For applying such a compression via low-rank decomposition, we used the newly suggested approach of~\cite{Liebenwein2021Compressing}, namely ALDS: Automatic Layer-wise Decomposition Selector. Given a specific compression ratio and a large neural network we wish to compress, ALDS aims at finding the optimal small (decomposed) architecture, by automatically identifying the optimal per-layer compression ratio, while simultaneously obtaining the desired overall compression. We utilize this approach in order to save time in searching for many hyper-parameters, such as the per-layer compression ratio.

\subsection{The Searching Algorithm}

To find the optimal combination between the defined options above, we apply an efficient search algorithm, which we call ``automatic compressed network finder'' (\textsc{ACNF}). 

\textbf{\textsc{ACNF}($\mathcal{N},\mathcal{F},\mathcal{C},k$)}. The input to our pipeline is (i) a set $\mathcal{N}$ of candidate pre-trained models (networks), (ii) a set $\mathcal{F}$ of candidate deep learning frameworks to use, (iii) a set $\mathcal{C}$ of candidate model compression methods, and  (iv) an integer $k\geq 1$, denoting the number of iterations.

\textbf{Initialization Step.} First, our algorithm initializes the following: a uniform sampling probability vector $u$ of size $|\mathcal{F}||\mathcal{N}||\mathcal{C}|$, where every entry in this vector corresponds to the probability of sampling a specific combination of a network, framework, and compression method. It also defines a table $T$ of $|\mathcal{F}||\mathcal{N}||\mathcal{C}|$ rows and $5$ columns. 

\textbf{Iterative execution.} Our Algorithm operates as follows during a period of $k$ iterations.

\begin{enumerate}[(i)]
    \item Sample a combination of a network $N$, a framework $F$, and a compression method $C$, based on the distribution vector $u$.
    \item Construct the network $N$ on the framework $F$.
    \item Compress the network $N$ via method $C$.
    \item Compute the accuracy (denoted by $acc$) or other required quality measure of the network and compute its running time on a single frame (denoted by $time$) on the desired edge device (RPI0 in our case).
    \item Save the tuple $(N,F,C,acc,time)$ in the global table $T$.
    \item \textsc{PairChecker}$(N,F,C)$: We now wish to learn from the results we obtained and not waste another iteration on an irrelevant sampled combination that will not yield improved results. In other words, if we observed that a pair from the tuple $(N,F,C)$ is repeatedly giving ``bad'' results, we exclude it from the sampling options (or at least reduce the probability of it being sampled). Bad results, in this case, can be defined according to the measure $m=acc/time$, where low values are bad, and high values are good. 
     We thus set a threshold $\alpha$, and for all the entries in $u$ that correspond to a pair from the tuple $(F,N,C)$, we multiply them by $m/alpha$. Note that, if $m$ is higher than the threshold we increase the probability of sampling each pair from $(N,F,C)$, while if $m$ is smaller than the threshold we reduce the same probability. Indeed we normalize the vector $u$ again to sum to $1$.
\end{enumerate}


\textbf{Observation.} Our algorithm significantly reduced the number of combinations we should check, as the probability of sampling a bad combination gets close to zero very fast. A clear example of that was the combination of the ALDS algorithm (see Section~\ref{seC:choices}) with the Tensorflow Light framework that was excluded very quickly from the sampling options, and we explain it as follows. The ALDS approach returns a compressed model, which has grouped convolution layers. This kind of layer (convolution) enables substantial memory savings (in terms of the number of parameters) with minimal loss of accuracy. Basically, it is done using a set of convolutions, i.e., multiple (parallel) kernels in a single layer - resulting in multiple channel outputs per layer, which leads to broader networks with the ability to learn a varied set of low-level and high-level features.
However, these layers are a very good example of how the choice of platform/framework and compression method must be synced, where these layers are highly optimized in TVM~\cite{9153227}, but their implementations in other modern deep learning frameworks are far from performing optimally in terms of speed. Thus our algorithm quickly excluded the combination of the ALDS algorithm with the Tensorflow Light framework from our search. 


\subsection{Comparison Table}

We now provide a table that compares different famous neural networks implemented above different deep learning platforms/frameworks and compressed by different methods. For each such combination, we report both the running time and the mean intersection over the Union (mIoU) metric, which is the ratio between the area of overlap and the area of union between the ground truth and the predicted areas (averaged across all images). This metric is known to be the best for measuring semantic segmentation quality. We ran our searching algorithm on images of size $513\times 513\times3$, and once the algorithm is done, we tested some of the good combinations on images of size $284\times284 \times3$. The results are given in Table~\ref{table:comparison}.


We note that in some cases, the compression adds to the running time instead of reducing it. This is due to the fact that compression may result in a new layer architecture such ass grouped convolutions, such layers may not be optimized on different frameworks.  These cases were not added to our table as they do not run faster nor more accurate than the original uncompressed model version.

\section{Our Low-Cost System}\label{sec:systems}

Being able to have a fast inference time semantic segmentation network, allows us to suggest several complicated systems that run efficiently on weak and cheap hardware devices.
In this section, we give our suggested system with  three different applications (many others can be added). Our systems are based on the following setup.

\textbf{Hardware. }The setup of the system consists of a DJI's Tello drone that carries a micro-computer (the Raspberry Pi Zero) that we connected on board via hot glue; see Figure~\ref{fig:tello}. The Raspberry Pi gets energy from the same battery of the Tello, which is one of the reasons that we used RPI0 and not the versions of RPI with higher computation power that requires a separate battery due to voltage issues. The Raspberry Pi is connected to a monocular RGB Pi Camera, which is the standard camera of the RPI. The Raspberry Pi controls the Tello via WIFI commands that imitate the remote controller of the Tello.

\textbf{Software. }We used the official software library (SDK) to control the Tello from the Rpi in real-time. Our code is written in Python 3.8~\cite{10.5555/1593511}, and after testing the packages PyTorch~\cite{paszke2019pytorch} TVM~\cite{chen2018tvm}, TensorFlow Lite~\cite{abadi2016tensorflow}, with several networks and several compressions; see Table~\ref{table:comparison}, we choose to use the LRASPP-MobileNetV3-Large with a $45\%$ Compression using the ALDS Algorithm, implemented on top of the Apache TVM framework, as it is among the most efficient solutions for our system with a reasonable accuracy/mIoU (mean intersection over union).

\textbf{Swap. } One of the main limitations of the Raspberry Pi Zero V2 is its very small RAM memory size, where it only has $512$MB of SDRAM memory, and for that reason, some more resource-demanding neural networks or frameworks, simply can not run due to insufficient memory capacity.
Hence, we used a swap file of $2048$MB in addition to the SDRAM memory, which is obviously much slower than the SDRAM on-board, but makes it possible to run these networks which we could not use without it.

\textbf{Led light. }Another feature that we added to our system, is a LED light that is connected to the Raspberry Pi. Users can use it generally for debug purposes, e.g., when one wants to make sure that an object they wanted to detect is properly detected. The light turns on when the desired object is detected by the network; See Figure~\ref{fig:bomb}.


\subsection{System \#1: Searching for people}
 \begin{figure}[t]
      \centering
      \includegraphics[width=8cm, height=2cm]{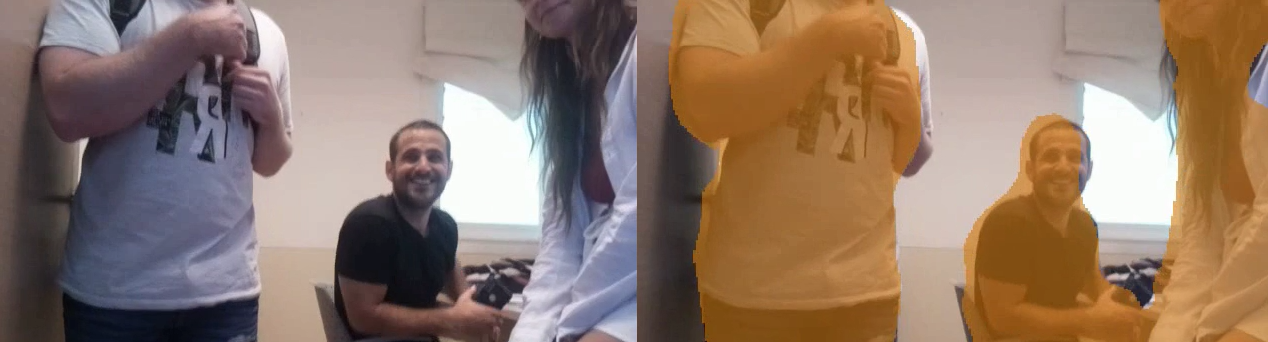}
      \caption{\textbf{Left:} The original video seen by the drone. \textbf{Right:} The semantic segmentation results to detect people}
      \label{fig:fire}
   \end{figure}
In this application, we aim at suggesting a semantic segmentation-based low-cost system for detecting people. It can be use for helping firefighters detect survivors in a building on fire, to be able to find them quickly and thus save their life.

We use the setup defined at the beginning of Section~\ref{sec:systems}

\textbf{Navigation. }For simplicity, when the system is turned on, the drone begins to navigate in a pre-defined path. Recent result~\cite{jubran2022newton} suggests SLAM (Simultaneous Localization and Mapping) for the Tello's drone, which may easily be combined with our system in future versions.

\textbf{Workflow.} During the drone navigation, the camera streams video to the Raspberry Pi, and our code process it at an average rate of $\sim5$ frames per second for a resolution of $284\times284$ RGB pixels. We also tested the system on frames with about doubled resolution of $513\times513$ with a speed trade-off of $2$ frames per second. 

Each frame is provided as input to our compressed network for semantic segmentation. The output is a matrix, which contains the predicted class number for each pixel in this frame. The class can be one of the pre-defined classes that the network was trained for, or ``unknown" if none of the classes was detected. For example, the companion network for this paper detects $20$ classes, including: ``person", ``table'', ``cat", ``dog" and ``sofa".

We refer the reader to Fig.~\ref{fig:fire} for illustration, and to the video in the supplemental material which has a fully running example. We also note that such a system can be used in many other applications such as detecting a thief in a building.

Note that we could use the LRASPP-MobileNetV3-Small to obtain an average rate of $\sim 10$ frames per second for a resolution of $284\times284$ RGB pixels, or a compressed version of it to make it even faster, however, we aim to show that even the used network (which is even not the best we found) is sufficient enough for these applications.  

\subsection{System \#2: Low-cost Parking Spot Detector}

Here we use the semantic segmentation to identifty cars and empty parking spots. It could be used to scan a parking place via a tiny drone to find an empty parking spot.
Our vision is that each user that comes to the parking area, presses on a bottom that invokes a drone that starts searching for a parking spot, once it finds one, the drone notifies the system and reserves this spot for the user who asked for it. 
For other users that come after, another drone is invoked, to scan the same path. Thus whoever comes first receives his parking spot first. 

\textbf{Workflow. } The drone begins to navigate and scan a pre-defined path, passing frames of images of size $m\times m$ for candidate parking spot places to our efficient network ($m\in \{284, 513\}$). The network process these images, and returns its output semantic segmentation matrix (per-pixel detected object). This matrix is then passed into an algorithm, which identifies based on the space which does not have a "car" label if there is a car that is parking in the spot or if it is empty; see Figure~\ref{fig:car} for illustration, and the example video in the supplemental material.


\subsection{System \#3: Suspicious Objects}
Here we aim at detecting a suspicious object reported by someone or a bomb. The drone scans the reported area searching for this object as previously done. See Figure~\ref{fig:bomb} for illustration, and the example video in the supplemental material. In the video and the Figure, we tread the bottle as a suspicious object or a bomb.

 \begin{figure}[t]
      \centering
      \includegraphics[scale=0.3]{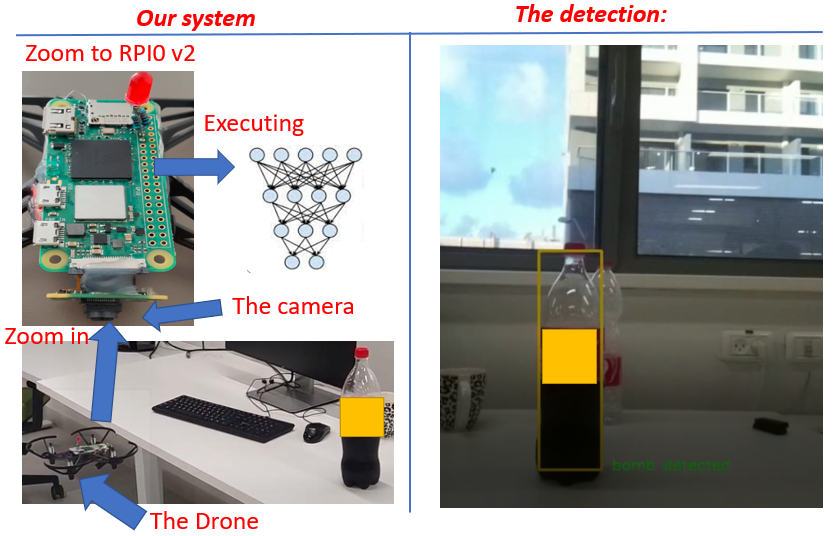}
      \caption{Left: Illustration of our system. Right: The applied detection.}
      \label{fig:bomb}
   \end{figure}

\section{CONCLUSIONS}

We presented the first system that runs real-time semantic segmentation via deep-learning on very weak, low-cost, low-weight, and low-powered IoT micro-computers such as the RPi0. The main challenge was to design the very specific right combination of networks, platforms, and compression techniques. We solved this by suggesting a searching algorithm that samples wisely a given combination to test. 

As an example application, we attached the RPi0 to a Tello DJI toy-drone that turned it into an autonomous drone that detects people, objects (indoor), or parking lots (outdoor) in real-time.

We expect that the provided open source and instructions will open the door for many future applications in robotics and IoT. Example of future work includes applying the system on: wearable devices, face recognition, simultaneous localization and mapping (SLAM), real-world products for security forces, firefighters, and commercial applications such as malls and deliveries.

\addtolength{\textheight}{-12cm}   









\bibliographystyle{apalike}
\bibliography{mybib}

\end{document}